\documentclass[lettersize,journal]{IEEEtran}
\usepackage{amsmath,amsfonts}
\usepackage{algorithm}
\usepackage{array}
\usepackage[caption=false,font=normalsize,labelfont=sf,textfont=sf]{subfig}
\usepackage{textcomp}
\usepackage{stfloats}
\usepackage{url}
\usepackage{verbatim}
\usepackage{graphicx}
\usepackage{cite}
\usepackage{color}
\usepackage{xcolor}
\usepackage{booktabs}
\usepackage{multirow}
\usepackage{enumitem}
\usepackage{algpseudocode}
\newcommand{\revmh}[1]{\textcolor{black}{#1}}
\newcommand{\attmh}[1]{\textcolor{black}{#1}}
\newcommand{\wsj}[1]{\textcolor{black}{#1}}

\begin{document}

\title{Vehicle-as-Prompt: A Unified Deep Reinforcement Learning Framework for Heterogeneous Fleet Vehicle Routing Problem}

\author{Shihong Huang, Shengjie Wang, Lei Gao, Hong Ma, Zhanluo Zhang, Feng Zhang, Weihua Zhou

\thanks{Shihong Huang, Shengjie Wang, and Hong Ma are with the Polytechnic Institute, Zhejiang University, 310015, Hangzhou, China}
\thanks{Lei Gao, Zhanluo Zhang, and Feng Zhang are with S.F. Technology Co., Ltd., 518054, Shenzhen, China}
\thanks{Weihua Zhou is with the School of Management, Zhejiang University, 310058, Hangzhou, China}
%\thanks{Corresponding author: Hong Ma, hongma@zju.edu.cn}
}

% \author[a1,a2]{Shihong Huang}
% \author[a1,a2]{Shengjie Wang}
% \author[a2]{Lei Gao}
% \author[a1]{Hong Ma\corref{cor1}}\ead{hongma@zju.edu.cn}
% \author[a2]{Zhanluo Zhang}
% \author[a2]{Feng Zhang}
% \author[a3]{Weihua Zhou}

% \address[a1]{Polytechnic Institute, Zhejiang University, 310058, Hangzhou, China}
% \address[a2]{S.F. Technology Co., Ltd., 518054, Shenzhen, China}
% \address[a3]{School of Management, Zhejiang University, 310058, Hangzhou, China}
% \cortext[cor1]{Corresponding author}

% % The paper headers
% \markboth{Journal of \LaTeX\ Class Files,~Vol.~14, No.~8, August~2021}%
% {Shell \MakeLowercase{\textit{et al.}}: A Sample Article Using IEEEtran.cls for IEEE Journals}

% \IEEEpubid{0000--0000/00\$00.00~\copyright~2021 IEEE}
% % Remember, if you use this you must call \IEEEpubidadjcol in the second
% % column for its text to clear the IEEEpubid mark.

\maketitle

\begin{abstract}
Unlike traditional homogeneous routing problems, the Heterogeneous Fleet Vehicle Routing Problem (HFVRP) involves heterogeneous fixed costs, \wsj{variable travel costs}, and capacity constraints, rendering solution quality highly sensitive to vehicle selection. Furthermore, real-world logistics applications often impose additional complex constraints, markedly increasing computational complexity. However, most existing Deep Reinforcement Learning (DRL)-based methods are restricted to homogeneous scenarios, leading to suboptimal performance when applied to HFVRP and its complex variants. To bridge this gap, we investigate HFVRP under complex constraints and develop a unified DRL framework capable of solving the problem across various variant settings. We introduce the Vehicle-as-Prompt (VaP) mechanism, which formulates the problem as a \wsj{single-stage autoregressive decision process}. Building on this, we propose VaP-CSMV, a framework featuring a cross-semantic encoder and a multi-view decoder that effectively addresses various problem variants and captures the complex mapping relationships between vehicle heterogeneity and customer node attributes. Extensive experimental results demonstrate that VaP-CSMV significantly outperforms existing state-of-the-art DRL-based neural solvers and achieves competitive solution quality compared to traditional heuristic solvers, while reducing inference time to mere seconds. Furthermore, the framework exhibits strong zero-shot generalization capabilities on large-scale and previously unseen problem variants, while ablation studies validate the vital contribution of each component. 
%\attmh{(shorten before submission to TRE.)}
\end{abstract}


\section{Introduction}
\IEEEPARstart{T}{he} Vehicle Routing Problem (VRP), a classical problem in combinatorial optimization, has extensive applications in logistics distribution, urban traffic scheduling, and supply chain management~\cite{schrijver2005history, yang2020goods, li2024mathematical, li2025collaborative}. Driven by the rapid growth of e-commerce and on-demand delivery, the heterogeneity of vehicle fleets has become increasingly pronounced, rendering the traditional homogeneous-fleet assumption inadequate for modeling real-world logistics operations. The Heterogeneous Fleet Vehicle Routing Problem (HFVRP) better reflects these distribution networks, enabling more effective optimization of operational costs and efficiency~\cite{kocc2015hybrid, li2022deep}. However, navigating its exponentially large, NP-hard search space remains a significant computational challenge. Traditional exact algorithms \cite{lawler1966branch} demand prohibitive computation times, while metaheuristic algorithms~\cite{li2010adaptive} often struggle with scalability and frequently converge to local optima, thereby limiting their applicability in large-scale or real-time routing scenarios. In recent years, Deep Reinforcement Learning (DRL) has emerged as a promising alternative for combinatorial optimization, leveraging its robust feature representation capabilities and \wsj{end-to-end learning method}. Attention-based models~\cite{kool2018attention, kwon2020pomo} have achieved solution qualities highly competitive with, and sometimes surpassing, traditional solvers on the Capacitated Vehicle Routing Problem (CVRP). Nevertheless, the vast majority of existing DRL research focuses exclusively on homogeneous fleets. In these conventional DRL frameworks, models primarily learn spatial dependencies within the topological space, treating vehicle state information merely as secondary, auxiliary features. Consequently, research extending DRL methodologies to HFVRP remains relatively scarce.

Existing DRL methods primarily face three major challenges when solving HFVRP.
\revmh{First, vehicle heterogeneity complicates the decision-making process.} While recent DRL studies address heterogeneous routing \cite{li2022deep, berto2024parco, wu2025efficient}, they predominantly focus on the Heterogeneous Capacitated Vehicle Routing Problem (HCVRP), which considers only capacity differences. Real-world applications, however, also involve varying fixed costs, variable travel costs, and fleet composition. These complexities amplify the impact of \revmh{vehicle dispatching}, as poor dispatching choices can easily offset the gains achieved from routing optimization. This necessitates the joint optimization of vehicle dispatching and routing, which remains underexplored in current DRL literature.
Second, current problem modeling approaches lack broad applicability~\cite{wu2024neural}. In practice, heterogeneous fleet composition rarely exists in isolation; they typically co-occur with constrained variants such as open routes, backhauls, distance limits, and time windows. Because training dedicated models for each variant incurs prohibitive computational overhead, efficient industrial deployment necessitates a unified foundational framework that supports multi-variant joint solving and scenario-specific fine-tuning~\cite{berto2024routefinder}. However, a unified DRL \wsj{framework} capable of integrating heterogeneous fleet composition with diverse VRP variants remains unexplored.
Third, the capabilities of existing solution methods are limited. Current methods for solving HCVRP fall into two main categories: the first adopts a hierarchical decision-making mechanism utilizing separate vehicle and node selection networks~\cite{li2022deep}, while the second introduces Multi-dimensional Pointer Networks~\cite{liu20242d} or Multi-Agent approaches~\cite{berto2024parco} for parallel, joint decision-making across multiple vehicles. However, the former approach often fails to integrate critical information regarding the relationship between heterogeneous vehicle attributes and customer node topology, which impairs the model's exploration capabilities. The latter allows all vehicles to participate simultaneously in decision-making at every time step; consequently, as the fleet size increases, \wsj{the memory requirements surge exponentially}, severely limiting practical scalability.

To address these challenges, we propose a novel \wsj{DRL} framework to efficiently solve HFVRP and its complex variants. Our main contributions are as follows. First, we propose a unified DRL framework capable of solving HFVRP and multiple variants, modeling multi-dimensional vehicle heterogeneity (capacities, fixed and variable travel cost, and vehicle availability). Second, we introduce the Vehicle-as-Prompt (VaP) mechanism, which streamlines joint vehicle and customer node selection into a single-stage autoregressive decision process, significantly reducing computational cost. \wsj{Third, we design a cross-semantic encoder and a multi-view decoder within the DRL framework}. By leveraging a \wsj{dual-attention} mechanism and cross-semantic feature fusion, the model seamlessly adapts to various HFVRP variants and captures the critical relationships among vehicle heterogeneity, customer node topology, and other constraints. Finally, extensive experiments demonstrate that VaP-CSMV achieves excellent solution quality across five fundamental variants, exhibiting robust zero-shot flexibility across diverse problem scales and unseen variants, while ablation studies validate the efficacy of each component.

\section{Related Work} \label{Literature}
\revmh{This section first provides a concise review of traditional methods for solving HFVRP. Subsequently, we explore recent advancements in DRL for VRPs, focusing on their evolving capability to handle heterogeneous fleets.}

\subsection{Traditional Methods for Solving HFVRP}
\revmh{Traditional approaches to VRP, HFVRP, and their variants primarily include exact, approximation, and heuristic algorithms. Research on HFVRP dates back to the seminal work of~\cite{golden1984fleet}, which introduced the Fleet Size and Mix (FSM) problem. This foundation was expanded by~\cite{taillard1999heuristic}, who formally defined HFVRP, thereby distinguishing strategic fleet composition from tactical routing decisions. Recent advancements in exact algorithms have increasingly integrated complex operational attributes. For instance, ~\cite{han2024formulations} developed branch-and-cut algorithms for HFVRP with soft time deadlines, while~\cite{yu2019branch} proposed a branch-and-price framework for green HFVRP with time windows. Despite these breakthroughs, exact methods remain computationally intractable for instances exceeding 100–200 customers, particularly when incorporating non-linear energy consumption or dynamic traffic patterns. Although these algorithms theoretically guarantee optimality, their prohibitive execution times severely restrict their utility in real-time applications. To overcome these efficiency bottlenecks, metaheuristic algorithms have emerged as the dominant mechanism for large-scale HFVRP variants. Notable examples include the multistart adaptive memory programming metaheuristic for HFVRP proposed by~\cite{li2010adaptive}, and the hybrid population heuristic for variants involving both fixed and variable travel cost developed by~\cite{liu2013hybrid}. Furthermore,~\cite{lai2016tabu} introduced a tabu search heuristic to address HFVRP on multigraphs, effectively accounting for diverse road network characteristics. Other robust frameworks include hybrid Iterated Local Search ~\cite{subramanian2012hybrid}, hybrid evolutionary algorithms ~\cite{kocc2015hybrid}, and, more recently, Adaptive Large Neighborhood Search (ALNS) for mixed-fleet electric vehicle routing~\cite{donmez2022mixed}. Nevertheless, despite their maturity and capacity for high-quality solutions, traditional metaheuristics rely heavily on hand-crafted rules and problem-specific designs. This dependency results in protracted development cycles and limited adaptability across varying scenarios. Such constraints, coupled with slow convergence on extremely large-scale problems, have catalyzed a research shift toward \wsj{DRL} methods to achieve rapid, robust, and end-to-end routing optimization.}

\subsection{DRL-based Methods for Solving VRP}
In recent years, DRL-based methods for solving combinatorial optimization problems have emerged as a prominent research focus. As a foundational step, ~\cite{vinyals2015pointer} proposed the Pointer Network, utilizing a sequence-to-sequence architecture and an attention mechanism to solve the Traveling Salesman Problem (TSP). Building upon this within the DRL framework, ~\cite{bello2016neural} introduced an actor-critic method to train these networks, significantly improving solution performance. A seminal contribution was subsequently made by~\cite{kool2018attention}, who proposed an attention model based on the transformer architecture. This model achieved efficiency approaching that of traditional heuristic methods on CVRP and has since become a widely adopted benchmark framework
for subsequent research. To further enhance performance,~\cite{kwon2020pomo} proposed Policy Optimization with Multiple Optima (POMO), which leverages the symmetry of the solution space and multi-start trajectory augmentation to substantially improve training stability and solution quality. Building on this,~\cite{kim2022sym} introduced symmetry-breaking techniques, that further optimized CVRP solution performance and significantly outperformed traditional ILS solvers on the prize-collecting TSP, achieving speedups of approximately 240 times. Subsequently,~\cite{berto2024routefinder} developed a unified neural solving framework for multiple VRP variants, thereby advancing the flexibility of neural solvers for VRP. 

Despite these advancements, research on \revmh{learning-based methods for HFVRP and its complex variants remains limited. Although~\cite{li2022deep} explored solving HCVRP, their approach primarily relies on a basic embedding of heterogeneous capacities, thereby neglecting other critical vehicle-specific attributes, such as fixed and 
variable travel costs.} Subsequent related works \cite{berto2024parco, liu20242d, hua2025camp, wu2025efficient}, while extending this approach, similarly address only HCVRP. They have not yet covered more practically significant heterogeneous variants with complex constraints, such as open routes, time windows, and various linehaul and backhaul requirements. Because these complex, constrained heterogeneous fleet problems are highly prevalent in real-world applications, developing a unified neural solver for a wide range of heterogeneous scenarios holds significant research value and practical importance.

\section{Problem Formulation}
In this section, we first introduce the mathematical formulation of \wsj{HFVRP} and subsequently present problem variants with additional constraints.
We then utilize the VaP mechanism to uniformly model these diverse variants as a Markov Decision Process (MDP).

\subsection{Mathematical Formulation}
\wsj{HFVRP} is defined on a complete directed graph $\mathcal{G}=(\mathcal{V}, \mathcal{E})$, comprising a node set $\mathcal{V}=\{0\} \cup \mathcal{C}$ (where the depot is indexed as $0$ and the customer set is $\mathcal{C}=\{1, \dots, N\}$) and an edge set $\mathcal{E}=\{(i,j) \mid i,j \in \mathcal{V}, i \neq j\}$. 
Each node $i$ in the graph has attributes such as coordinates $loc_i$ and linehaul (delivery) demand $q_i^L$. The travel distance of edge $(i,j)$ is denoted by $c_{ij}$. Each vehicle $k$ in the heterogeneous fleet $\mathcal{K}$ has a maximum capacity $Q_k$, a fixed cost $fc_k$, and a \wsj{variable travel cost} $ac_k$. To represent the fleet trajectories and the flow of goods, the model introduces two sets of decision variables:  binary variables $x_{ijk} \in \{0, 1\}$, which equals $1$ if vehicle $k$ travels directly from node $i$ to node $j$, and $0$ otherwise; and non-negative continuous variables $u_{ik} \geq 0$, representing the remaining delivery load of vehicle $k$ upon leaving node $i$. The following Mixed Integer Linear Programming (MILP) model is formulated to minimize the total operational cost:

\begin{align}
\text{Minimize } & Z = \sum_{k \in \mathcal{K}} \left( fc_k \sum_{j \in \mathcal{C}} x_{0jk} + ac_k \sum_{(i,j) \in \mathcal{E}} {c}_{ij} x_{ijk} \right) \label{eq:objective} \\[1em]
\text{s.t.} & \sum_{k \in \mathcal{K}} \sum_{j \in \mathcal{V}, j \neq i} x_{ijk} = 1, \quad \forall i \in \mathcal{C} \label{eq:visit} \\
& \sum_{j \in \mathcal{V}, j \neq i} x_{ijk} = \sum_{j \in \mathcal{V}, j \neq i} x_{jik}, \quad \forall i \in \mathcal{C}, \forall k \in \mathcal{K} \label{eq:flow} \\
& \sum_{j \in \mathcal{C}} x_{0jk} \leq 1, \quad \forall k \in \mathcal{K} \label{eq:fleet_limit} \\
& \begin{aligned}
    & u_{jk} \leq u_{ik} - q_j^L x_{ijk} + M(1-x_{ijk}), \\
    & \quad \forall i \in \mathcal{V}, j \in \mathcal{C}, k \in \mathcal{K} 
  \end{aligned} \label{eq:cap_linehaul} \\[0.5em]
& u_{ik} \leq Q_k, \quad \forall i \in \mathcal{V}, k \in \mathcal{K} \label{eq:cap_hetero} \\
& x_{ijk} \in \{0, 1\}, \quad \forall i \in \mathcal{V}, j \in \mathcal{V}, i\neq j, \forall k \in \mathcal{K} \label{eq:domain_x} \\
& u_{ik} \geq 0 , \quad \forall i \in \mathcal{V}, \forall k \in \mathcal{K} \label{eq:domain_u}
\end{align}

The objective function (\ref{eq:objective}) minimizes the sum of vehicle fixed costs and variable travel costs. Constraints (\ref{eq:visit}) guarantee that each customer is served exactly once. Constraints (\ref{eq:flow}) ensure flow conservation and route continuity. Constraints (\ref{eq:fleet_limit}) ensure that each vehicle is deployed at most once. Constraints (\ref{eq:cap_linehaul}) track the progressive decrease of the delivery load along the route and eliminate subtours. Constraints (\ref{eq:cap_hetero}) ensure that the total load of any vehicle does not exceed its maximum capacity. Finally, Constraints (\ref{eq:domain_x}) and (\ref{eq:domain_u}) define the domains of the decision variables.

\begin{figure}[!t]
\centering
\includegraphics[width=0.9\columnwidth]{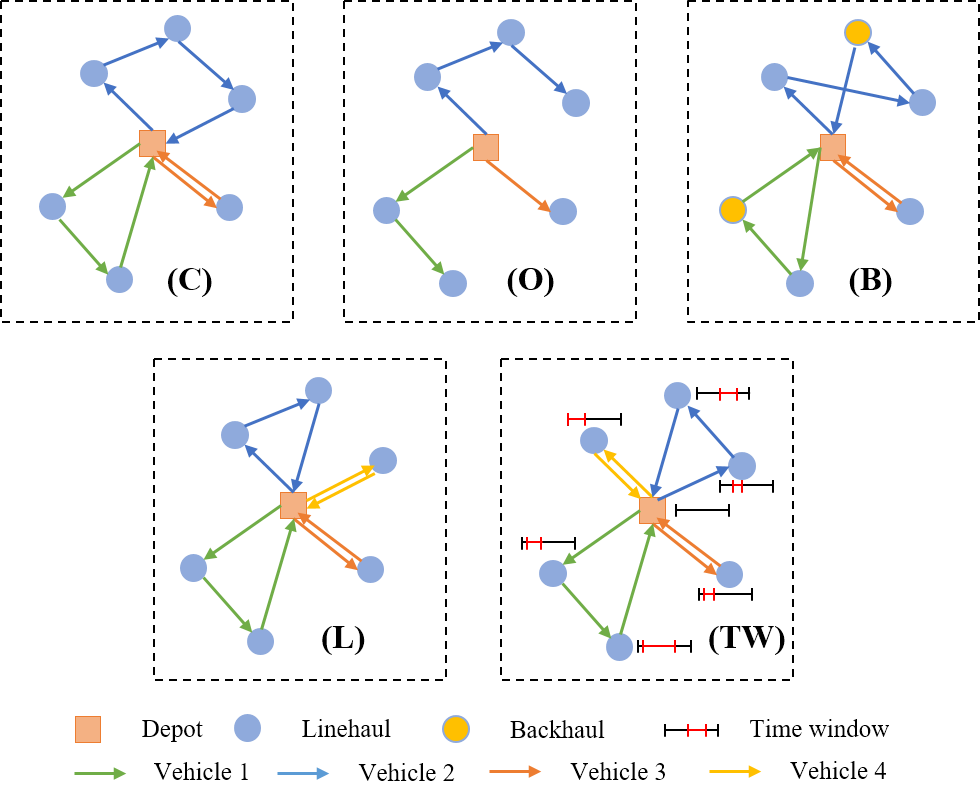} % 
\caption{\wsj{Illustration of the five HFVRP variants and their associated practical constraints. These constraints are categorized into node-level \revmh{requirements}, including Capacity (C), Backhauls (B), and Time Windows (TW), and route-level \revmh{requirements}, including Open routes (O) and Distance limits (L).}}
\label{fig:problems}
\end{figure}

Building upon HFVRP, we consider five variants that incorporate diverse practical constraints~\cite{zhou2024mvmoe}, as illustrated in Figure~\ref{fig:problems}:
\begin{enumerate}
\item \textbf{Capacity (C)}: Each customer node $i \in \mathcal{C}$ is associated with a specific delivery demand $q_i^L$. 
\item \textbf{Open routes (O)}: Vehicles are not required to return to the depot after visiting their assigned customers\wsj{.} 
\item \textbf{Backhauls (B)}: While standard \wsj{HFVRP} typically considers only delivery demands, practical applications often require vehicles to also pick up goods at customer nodes, known as backhaul demands. Consequently, \wsj{HFVRP} with backhauls must account for both delivery and pickup operations. In this work, we impose a strict precedence constraint on these operations: all delivery tasks (linehauls) must be completed before any pickup tasks (backhauls) begin.
\item \textbf{Distance limits (L)}: To maintain reasonable driver workloads, the total distance of each route must not exceed a specified upper bound.
\item \textbf{Time windows (TW)}: Each customer node $i \in \mathcal{C}$ has a specified time window $[e_i, l_i]$ and an associated service duration $s_i$. A vehicle must begin serving customer $i$ within the time interval $[e_i, l_i]$. If a vehicle arrives before $e_i$, it must wait until time $e_i$ to begin service. Furthermore, all vehicles must return to the depot no later than its closing time $l_0$. 
\end{enumerate}
% These four variants are illustrated in Figure~\ref{fig:problems}.

\subsection{Reformulation as Markov Decision Process}
\begin{figure}[!t]
\centering
\includegraphics[width=\columnwidth]{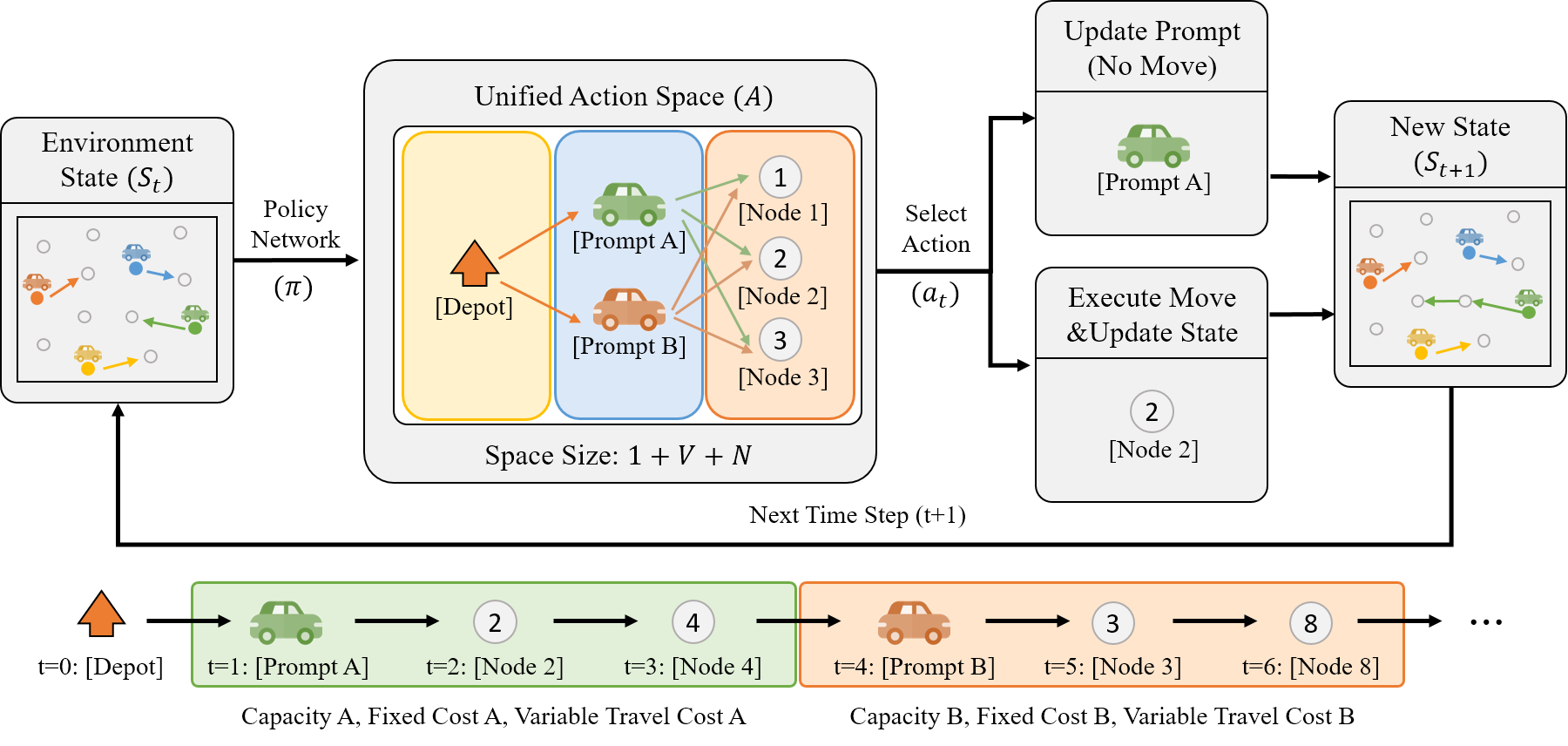} % 
\caption{\textbf{Detailed illustration of the Vehicle-as-Prompt (VaP) mechanism.} By abstracting heterogeneous vehicles as prompts, the VaP mechanism projects the depot, $V$ vehicle types, and $N$ customer nodes into a unified action space of size $1 + V + N$. updates the contextual constraints (e.g., capacity, travel costs, etc.). %without inducing physical movement. 
illustrates the transformation of HFVRP into an \revmh{autoregressive decision process.}}
\label{fig:vap}
\end{figure}

%To solve the heterogeneous vehicle routing problem, this paper draws inspiration from the tokenization techniques of large language models to map heterogeneous vehicles and customer nodes into a unified action space, thereby constructing the VaP mechanism. (Figure \ref{fig:vap} details the VaP mechanism.) Our VaP mechanism is primarily responsible for mapping heterogeneous vehicles and customer nodes into the same action space. 
To solve HFVRP, this paper draws inspiration from the tokenization techniques of large language models (LLMs) to construct the VaP mechanism. As detailed in Figure~\ref{fig:vap}, this mechanism effectively maps both heterogeneous vehicles and customer nodes into a unified action space. Selecting a customer node as an action corresponds to an actual physical movement of the vehicle, which subsequently updates the load, distance, and time states. On the other hand, selecting a vehicle type is treated as introducing a specific prompt; this action generates no physical displacement but establishes the operational constraints (i.e., vehicle capacity $Q_k$ and cost parameters $fc_k, ac_k$) for the subsequent route. Consequently, the routing problem can be reformulated as a single-stage autoregressive sequence generation task, where the corresponding MDP is defined by the tuple $\mathcal{M} = \langle \mathcal{S}, \mathcal{A}, \mathcal{T}, \mathcal{R} \rangle$:

%Selecting a customer node action represents an actual physical movement of the vehicle, correspondingly updating the load, distance, and time states. Selecting a vehicle type action is treated as introducing a specific prompt; this action generates no physical displacement but establishes constraint boundaries (i.e., vehicle capacity $Q_k$ and cost parameters $fc_k, ac_k$) for the subsequently generated route. Consequently, the problem can be reformulated as an autoregressive sequence, and its corresponding MDP is defined by a tuple $\mathcal{M} = \langle \mathcal{S}, \mathcal{A}, \mathcal{T}, \mathcal{R} \rangle$:

\textbf{State Space $\mathcal{S}$}: At time step $t$, the environment state $s_t$ comprises both static features and a dynamic context. The static features include the \revmh{problem feature vector $v_p = [\varphi_o, \varphi_b, \varphi_l, \varphi_{tw}] \in \mathbb{R}^{D_p}$}, which controls the problem variants; the coordinates $loc_i$, linehaul and backhaul demands $q_i^L, q_i^B$, hard time windows $[e_i, l_i]$, and service duration $s_i$ for each node; and the physical and economic attributes of each vehicle (maximum capacity $Q_k$, fixed cost $fc_k$, and variable travel cost $ac_k$). The dynamic context encompasses the currently \revmh{active} vehicle type, its current node location, the remaining available quantity of each vehicle type, and the \revmh{active} %current 
vehicle's cumulative travel distance, time, and load.  
Furthermore, it includes a binary visit status mask $v_{t,i}$ for each node, where $v_{t,i}=1$ indicates that the node has been visited.

\textbf{Action Space $\mathcal{A}$}: The discrete action space is defined as $\mathcal{A} = \{0\} \cup \mathcal{K} \cup \mathcal{C}$. At each time step $t$, the agent selects a single action $a_t \in \mathcal{A}$. These actions fall into three categories: $a_t = 0$ denotes returning to the depot, thereby concluding the \revmh{active} vehicle's route; $a_t \in \mathcal{K}$ denotes selecting a specific vehicle type as a prompt; and $a_t \in \mathcal{C}$ denotes choosing to visit a customer node.

\textbf{State Transition Rules $\mathcal{T}$}: Within the VaP mechanism, state transitions must strictly satisfy the MILP constraints while adhering to the syntactic rules of sequence generation. To achieve this, we introduce a rule-based masking mechanism. Specifically, we define a constraint mask $M_t^{\text{cons}}(a_t)$. Assuming $k$ is the active vehicle prompt and $j$ is the candidate customer node corresponding to action $a_t$, the mask is set to $M_t^{\text{cons}}(a_t) = 1$ if and only if all of the following conditions are met:
\begin{itemize}
    \item \textbf{Accessibility constraint}: Customer node $j$ has not been previously visited;
    \item \textbf{Fleet size constraint}: The vehicle type associated with action $a_t$ has vehicles available for dispatch;
    \item \textbf{Capacity constraint (C)}: The vehicle's remaining capacity is sufficient to accommodate the demand of node $j$, ensuring the total vehicle load does not exceed its upper limit;
    \item \textbf{Time window constraint (TW)}: If $\varphi_{tw} = 1$, the arrival time of vehicle $k$ at node $j$ must satisfy the hard time window requirements, and the vehicle must be able to return to the depot before the closing time $l_0$;
    \item \textbf{Backhaul priority constraint (B)}: If $\varphi_{b} = 1$, all linehaul (delivery) tasks must be completed before any backhaul (pickup) tasks can commence;
    \item \textbf{Open route constraint (O)}: If $\varphi_{o}=1$, vehicles are not required to return to the depot upon completing their routes;
    \item \textbf{Distance limit constraint (L)}: If $\varphi_{l}=1$, visiting node $j$ must not cause vehicle $k$ to exceed its maximum travel distance limit.
\end{itemize}

Next, we define a %syntactic 
\revmh{mask $M_t^{\text{g}}(a_t)$} to enforce the valid generation order of tokens within the sequence. Let $i_t$ denote \attmh{the active physical or virtual node} corresponding to the current decision state. Formally, $M_t^{\text{g}}(a_t)$ is defined as follows:
$$
M_t^{\text{g}}(a_t) = 
\begin{cases}
1, & \text{if } i_t = 0, a_t \in \{k \in \mathcal{K} \} \\
1, & \text{if } i_t \in \mathcal{K} \cup \mathcal{C}, a_t \in \{0\} \cup \{j \in \mathcal{C} \mid v_{t,j} = 0\} \\
0, & \text{others}
\end{cases}
$$

This formulation ensures that if the agent is at the start of a new route ($i_t = 0$), it must select an available vehicle type as a prompt. \revmh{On the other hand, if a vehicle is already active ($i_t \in \mathcal{K} \cup \mathcal{C}$), the agent is restricted to selecting an unvisited customer node, or choosing $0$ to terminate the active vehicle's route.}

\textbf{Reward Function $\mathcal{R}$}: To align with the objective of minimizing the total operational cost, the single-step reward $r_t$ is defined as the negative incremental cost incurred at time step $t$. Because the action space unifies both vehicle types and physical nodes, the fixed costs and variable travel costs are naturally decoupled:
\begin{equation}
    r_t(s_t, a_t) = 
    \begin{cases}
    -fc_{a_t}, & \text{if } a_t \in \mathcal{K} \\
    -ac_{k_t} \cdot c_{i_t, a_t}, & \text{if } a_t \in \{0\} \cup \mathcal{C} \\
    0, & \text{otherwise}
    \end{cases}
\end{equation}

Here, $k_t$ denotes the active vehicle and $i_t$ represents its current location. Furthermore, unlike DRL methods for homogeneous VRPs, which typically assume an unlimited fleet size, the restricted vehicle count in a heterogeneous fleet can lead to infeasible states during the decoding process. In such scenarios, the agent may encounter states where unvisited nodes remain while no valid actions are available due to fleet exhaustion. Traditionally, such infeasible solutions are penalized with static constant values, which can frequently cause gradient instability during policy network training. To address this, we introduce a dynamic penalty mechanism based on the set of unvisited nodes $\mathcal{U}_t$. Upon detecting an infeasible state, \revmh{the decoding sequence} is forcibly terminated, and a penalty $R_{p}$ is applied. This penalty represents the maximum estimated cost of serving each remaining node independently via direct round trips from the depot:
\begin{equation}
    R_{p} = - \sum_{i \in \mathcal{U}_t} \left[ \max_{k \in \mathcal{K}}(ac_k) \cdot \left(c_{0i} + c_{i0}\right) + \max_{k \in \mathcal{K}}(fc_k) \right]
\end{equation}

\revmh{By assigning a physically meaningful cost to infeasible solutions, this mechanism avoids the use of arbitrary, extreme penalties. Consequently, it prevents gradient instability during the training of the policy network.}

\section{Methodology}
\begin{figure*}[t]
\centering
\includegraphics[width=1.8\columnwidth]{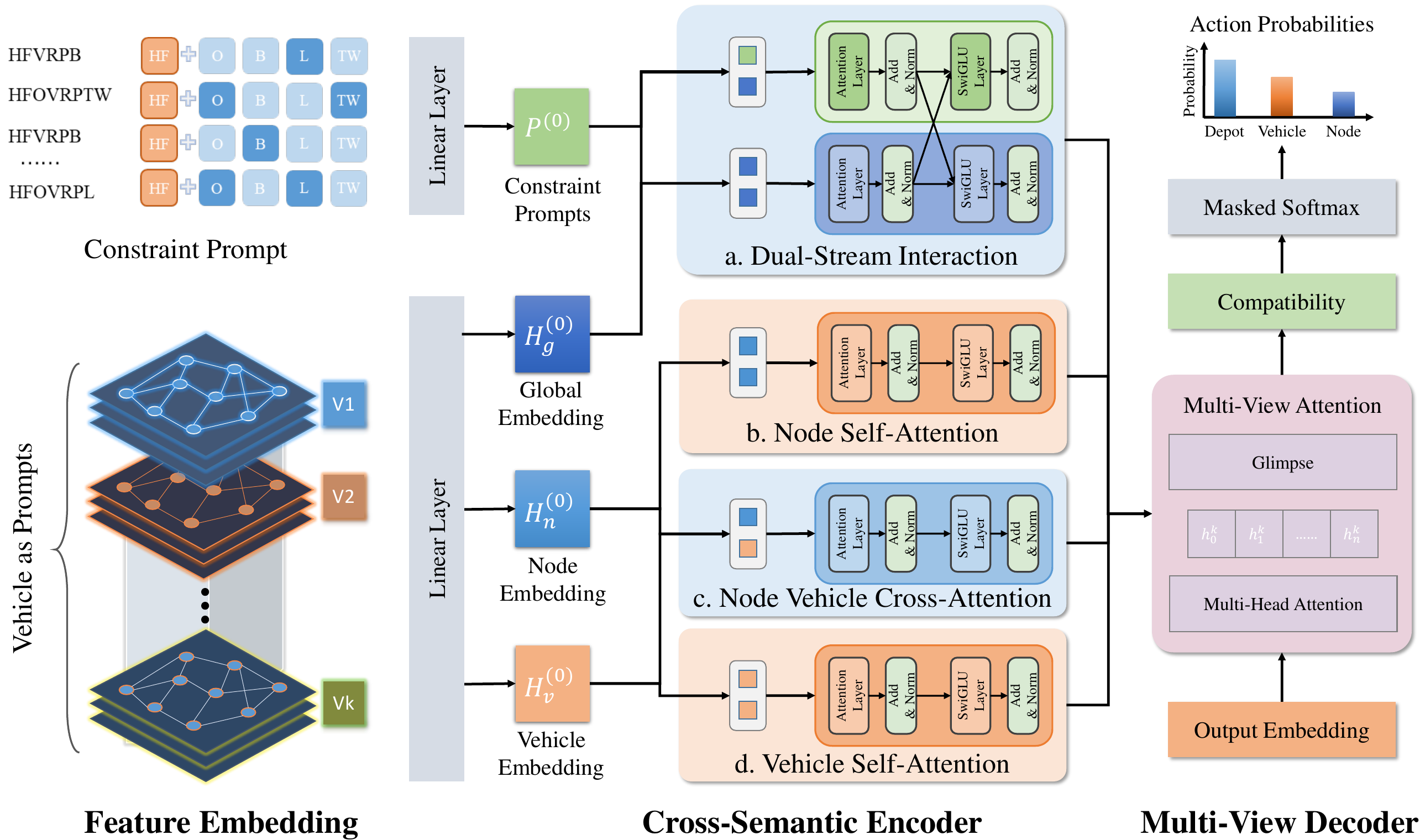} % 
\caption{\textbf{The overall architecture of VaP-CSMV.} The Feature Embedding Layer projects the environmental state and the problem \wsj{feature} into a unified high-dimensional embedding space. The Cross-Semantic Encoder extracts representations across four semantic domains—node-level, vehicle-level, vehicle-node interactions, and global contexts—outputting a comprehensive set of multi-semantic contextual \wsj{embeddings}. The Multi-View Decoder generates the probability distribution over the next node to visit, progressively constructing a feasible routing solution.}
\label{nn model}
\end{figure*}

To address the complex constraints inherent in HFVRP and effectively capture the relationships among diverse features, %within the VaP mechanism, 
we propose a novel DRL framework, termed VaP-CSMV, comprising three core modules: a Feature Embedding Layer, a Cross-Semantic Encoder, and a Multi-View Decoder. The overall architecture of VaP-CSMV is illustrated in Figure~\ref{nn model}.

\subsection{Feature Embedding}
First, the problem \revmh{feature vector} $v_p \in \mathbb{R}^{D_p}$, which encodes the specific variant type of the routing problem, is processed through a Multilayer Perceptron (MLP) followed by Layer Normalization (LayerNorm) to generate the initial constraint prompt embedding $P^{(0)}$.  \revmh{This corresponds to the Constraint Prompt module within Feature Embedding Layer (see Figure~\ref{nn model}).} This embedding serves as a global descriptor of the current problem configuration. Formally, this transformation is expressed as:
\begin{equation}
    P^{(0)} = \text{LayerNorm}(v_p W_a + b_a) W_b + b_b
\end{equation}
where $W_a \in \mathbb{R}^{D_p \times d_h}$ and $W_b \in \mathbb{R}^{d_h \times d_h}$ are trainable weight matrices, $b_a, b_b \in \mathbb{R}^{d_h}$ are trainable bias vectors, and $d_h$ denotes the hidden dimension.

Next, the model embeds the state of the routing environment. Within the MDP formulation, %this state comprises depot features $x^d$, customer node features $x^c$, and vehicle features $x^v$. Specifically, $x^d$ denotes the depot attributes, $x^c$ encompasses customer node attributes (e.g., coordinates, demands, service times, and time windows), and $x^v$ represents vehicle attributes (e.g., maximum capacity, fixed costs, and variable travel costs).  
this state comprises depot features $x^d$, customer node features $x^c$ (e.g., coordinates, demands, service times, and time windows), and vehicle features $x^v$ (e.g., maximum capacity, fixed costs, and variable travel costs).
To project these heterogeneous inputs into a shared embedding space, the model applies linear transformations to map them uniformly into $d_h$-dimensional embeddings. Additionally, the depot and vehicle features are concatenated and jointly projected to capture their coupled relationship, facilitating subsequent semantic interaction modeling. Formally, this state embedding process is expressed as:
\begin{align}
    & E^d = x^d W_d, \quad E^c = x^c W_c, \quad E^v = x^v W_v \\
    & E^{vd} = \text{Concat}(x^d, x^v) W_{vd}
\end{align}
where $W_d, W_c, W_v$, and $W_{vd}$ are trainable weight matrices. Building upon these embedded representations, the model constructs the initial embeddings for the respective semantic branches. Specifically, the node semantic branch aggregates depot and customer node features; the vehicle semantic branch combines depot and vehicle features; and the global semantic branch integrates all feature sets to provide a holistic view of the environment. Accordingly, the initial node embedding $H_n^{(0)}$, vehicle embedding $H_v^{(0)}$, joint vehicle-depot embedding $H_{vd}^{(0)}$, and global embedding $H_g^{(0)}$ are constructed as follows:
\begin{align}
    H_n^{(0)} &= [E^d, E^c] \\
    H_v^{(0)} &= [E^d, E^{vd}] \\
    H_{vd}^{(0)} &= [E^d, E^{vd}] \\
    H_g^{(0)} &= [E^d, E^{vd}, E^c]
\end{align}

\subsection{Cross-Semantic Encoder}
Following the feature embedding phase, the initial sequences are processed through a cross-semantic encoder. This encoder comprises multiple parallel attention blocks, with each block containing a Multi-Head Attention (MHA) mechanism, Layer Normalization (LayerNorm), and a SwiGLU-based non-linear feed-forward network. The SwiGLU mechanism is formalized as follows.
\begin{equation}
    \text{SwiGLU}(x) = (\text{SiLU}(x W_{f1}) \odot (x W_{f2})) W_{f3}
\end{equation}
where $\odot$ denotes element-wise multiplication, and $W_{f1}, W_{f2}$, and $W_{f3}$ are trainable \revmh{weight matrices}.

Within the cross-semantic encoder, the node and vehicle embeddings iteratively refine their representations via independent self-attention mechanisms. This corresponds to the Node Self-Attention and Vehicle Self-Attention modules within the Cross-Semantic Encoder (see Figure~\ref{nn model}). For the $l$-th layer, let $k \in \{n, v\}$ index the node and vehicle branches, respectively. The self-attention update for each branch is formulated as:
\begin{align}
    \hat{H}_k^{(l)} &= \text{LayerNorm} \Bigl( H_k^{(l-1)} + \text{MHA}(Q=H_k^{(l-1)}, \nonumber \\
    & \qquad K=H_k^{(l-1)}, V=H_k^{(l-1)}) \Bigr) \label{eq:mha_layer} \\[0.5em]
    H_k^{(l)} &= \text{LayerNorm} \bigl( \hat{H}_k^{(l)} + \text{SwiGLU}(\hat{H}_k^{(l)}) \bigr) \label{eq:ffn_layer}
\end{align}

To capture the dynamic mapping relationship between vehicle service capabilities and node demands, the model introduces a vehicle-node cross-attention mechanism. This corresponds to the Node-Vehicle Cross-Attention module in the Cross-Semantic Encoder (see Figure~\ref{nn model}). Specifically, this  mechanism utilizes the joint vehicle-depot embeddings \wsj{$H_{v}^{(l-1)}$} as the Query, and the complete node sequence $H_n^{(l-1)}$ as the Key and Value. Formally, this feature interaction is expressed as:
\begin{align}
    \hat{H}_{vd}^{(l)} &= \text{LayerNorm} \Bigl( H_{vd}^{(l-1)} + \text{MHA} \bigl( Q=H_{vd}^{(l-1)}, \nonumber \\
    & \qquad K=H_n^{(l-1)}, V=H_n^{(l-1)} \bigr) \Bigr) \label{eq:cross_attn} \\[0.8em]
    H_{vd}^{(l)} &= \text{LayerNorm} \bigl( \hat{H}_{vd}^{(l)} + \text{SwiGLU}(\hat{H}_{vd}^{(l)}) \bigr) \label{eq:ffn_output}
\end{align}
In addition to local feature interactions, the model introduces a dual-attention mechanism to integrate the global environmental context with the constraint prompts. This corresponds to the Dual-Attention module within the Cross-Semantic Encoder (see Figure \ref{nn model}). Each network layer incorporates Multi-Head Attention (MHA), a SwiGLU-based Feed-Forward Network (FFN), LayerNorm, and residual connections. Let the expanded sequence containing the constraint prompts and vehicle features be defined as $C^{(0)} = \text{Concat}[P^{(0)}, E^v]$. At the $l$-th layer, the global branch and the prompt-augmented branch are first updated in parallel and then exchange features via residual connections, enabling deep fusion between the routing state and the constraint prompts. Formally, this update process is expressed as:
\begin{align}
\hat{H}_g^{(l)}
= \mathrm{RMSNorm}^{(l)}\!\Big(
&H_g^{(l-1)}
+ \mathrm{MHA}^{(l)}\!\big(
H_g^{(l-1)}, \notag\\
&\mathrm{Concat}[H_g^{(l-1)}, C^{(l-1)}]
\big)
\Big) \\
\tilde{H}_g^{(l)}
= \mathrm{RMSNorm}^{(l)}\!\Big(
&\hat{H}_g^{(l)}
+ \mathrm{SwiGLU}^{(l)}\!\big(\hat{H}_g^{(l)}\big)
\Big) \\
\hat{C}^{(l)}
= \mathrm{RMSNorm}^{(l)}\!\Big(
&C^{(l-1)}
+ \mathrm{MHA}^{(l)}\!\big(
C^{(l-1)}, \notag\\
&\mathrm{Concat}[H_g^{(l-1)}, C^{(l-1)}]
\big)
\Big) \\
C^{(l)}
= \mathrm{RMSNorm}^{(l)}\!\Big(
&\hat{C}^{(l)}
+ \mathrm{SwiGLU}^{(l)}\!\big(\hat{C}^{(l)}\big)
\Big)
\end{align}

By leveraging this mechanism, the cross-semantic encoder produces a set of contextual representations that jointly encode vehicle, customer node, vehicle-node interaction, and global constraint features. These representations serve as the foundation for subsequent routing decisions.

\subsection{Multi-View Decoder}
Given the complex mapping relationships within the VaP mechanism, decoding from a single embedding view is prone to losing local details, potentially leading to suboptimal decisions in the action space. To address this, we propose a multi-view contextual fusion decoder that generates solutions sequentially in an autoregressive manner. The decoder first applies linear transformations to the four independent semantic embeddings (global $H_g$, node $H_n$, vehicle $H_v$, and vehicle-node interaction $H_{vd}$) output by the encoder, caching their Key and Value matrices. Let $k \in \{g, n, v, vd\}$ index distinct semantic views, $H^{t-1}_g$ denote the embedding of the current node, and $f_{\text{status}}^t$ represent the embedding of dynamic environmental features. These features encompass the active vehicle's remaining capacity, accumulated travel distance, and cumulative travel time, alongside the real-time availability of each vehicle type. The projection process is formulated as:
\begin{align}
    & K_k = H_k W_K^k, \\
    & V_k = H_k W_V^k, \\
    & Q^t = \text{Concat}(H^{t-1}_g, f_{status}^t) W_Q
\end{align}
where $W_K^k, W_V^k$, and $W_Q$ are weight matrices.
For each view $k$, the context vector $C_k^t$ is computed as:
\begin{equation}
    C_k^t = \text{Softmax}\left(\frac{Q^t (K_k)^T}{\sqrt{d_k}}\right) V_k
\end{equation}
where $d_k$ denotes the dimension of a single attention head. The model aggregates the four independent view contexts with equal weights and fuses them via a linear transformation:
\begin{equation}
    M^t = \left( \sum_{k \in \{g, n, v, vd\}} C_k^t \right) W_{\text{cmb}}
\end{equation}

Finally, the model employs a Pointer Network mechanism to calculate the similarity score $u_i^t$ between $M^t$ and the global \wsj{embedding} $H_g$. To ensure the solution's feasibility, it introduces a dynamic mask vector $\mathcal{M}^t$. If node $i$ violates any constraint at the current step, $\mathcal{M}_i^t = -\infty$; otherwise, $\mathcal{M}_i^t = 0$. The decoder ultimately outputs the probability distribution of the next node $i$ to visit via the Softmax function:
\wsj{
\begin{align}
    u_i^t &= M^t (H_g^{(i)})^T \\
    p(\pi_t = i \mid s_t) &= \frac{\exp(u_i^t + \mathcal{M}_i^t)}{\sum_{j=1}^{N_{all}} \exp(u_j^t + \mathcal{M}_j^t)}
\end{align}
}
Based on this probability distribution $p(\pi_t \mid s_t)$, the model employs a sampling strategy to select the next action, facilitating exploration of the state space during both the training and inference phases.

\subsection{Training Method}
In this study, we employ a REINFORCE-based training method\cite{williams1992simple}. To enhance training stability, we introduce a sample-based shared baseline strategy. To further improve the balance between exploration and exploitation, we incorporate a covariance-guided adaptive entropy control mechanism~\cite{ppo} and an entropy coefficient regularization method~\cite{entropy_cov} to guide the training process.

\subsubsection{Sample-Based Shared Baseline}
During training, we generate $S$ parallel trajectories $\tau_s^b$ for each instance $b$ within a batch $B$ using the policy $\pi_\theta$. The shared baseline $l(b)$ is defined as the average reward across these trajectories: $l(b) = \frac{1}{S} \sum_{s=1}^{S} R(\tau_s^b)$. By defining the advantage function as $A(\tau_s^b) = R(\tau_s^b) - l(b)$, the policy gradient for maximizing the expected reward $J(\theta)$ is approximated as:
\begin{equation}
    \nabla_\theta J(\theta) \approx \frac{1}{B \times S} \sum_{b=1}^{B} \sum_{s=1}^{S} A(\tau_s^b) \nabla_\theta \log \pi_\theta(\tau_s^b \mid b)
\end{equation}

\subsubsection{Entropy Coefficient Regularization}
We augment the objective function with an entropy regularization term $\sigma \mathcal{H}(\pi_\theta)$. By penalizing overconfident action distributions, this regularization mechanism maintains a baseline level of policy diversity throughout training.

\subsubsection{Covariance-Guided Entropy Control}
Standard regularization often struggles to prevent premature convergence in high-confidence regions. As demonstrated by~\cite{entropy_cov}, the change in policy entropy during parameter updates is governed by the covariance between the log-probability and the advantage function:
\begin{equation}
\begin{aligned}
    &\mathcal{H}(\pi_\theta^{t+1} \mid s) - H(\pi_\theta^{t} \mid s) \\
    &\approx -\beta \operatorname{Cov}_{a \sim \pi_\theta^{t}} \Bigl( \log \pi_\theta^{t}(a \mid s), \pi_\theta^{t}(a \mid s) \cdot A(s,a) \Bigr)
\end{aligned}
\end{equation}

Leveraging this property, we introduce a selective gradient-blocking mechanism to actively sustain diversity at critical decision nodes. Specifically, we define a dynamic mask $M_i$ to selectively filter actions with high covariance, incorporating a detachment probability $p_{\text{detach}}$:

\begin{equation}
    M_i = \mathbb{I}(\text{Cov}_i \geq \eta) \cdot \mathbb{I}(\alpha_i < p_{\text{detach}}) \quad \alpha_i \sim \mathcal{U}(0,1)
\end{equation}
where $\eta$ is a predefined threshold used to identify high-covariance actions, and $\mathbb{I}(\cdot)$ denotes the indicator function. For actions where the mask is activated ($M_i=1$), the corresponding gradients are detached during backpropagation. The final modified loss function $\mathcal{L}$ is formulated as:
\begin{equation}
    \mathcal{L} = \frac{1}{B \times S} \sum_{b=1}^{B} \sum_{s=1}^{S} A(\tau_s^b) \nabla_\theta \log \tilde{\pi}_\theta(\tau_s^b \mid b) + \sigma \mathcal{H}(\pi_\theta) 
\end{equation}
\revmh{where the modified log-probability $\log \tilde{\pi}_\theta$ is obtained by applying a stop-gradient operator to $\log \pi_\theta$ when $M_i=1$, while retaining the original gradient flow otherwise.} This mechanism explicitly suppresses gradient updates for high-covariance actions, preventing the model from becoming trapped in local optima.

\begin{table*}[t]
\centering
\caption{Detailed Performance Comparison on Datasets for Problem Size $N = 50$, $K = 20$.}
\label{tab:baseline_50}
\resizebox{\textwidth}{!}{
\begin{tabular}{lccccccccccccccc}
\toprule
\multirow{2}{*}{Method} & \multicolumn{3}{c}{HFCVRP} & \multicolumn{3}{c}{HFOVRP} & \multicolumn{3}{c}{HFVRPB} & \multicolumn{3}{c}{HFVRPL} & \multicolumn{3}{c}{HFVRPTW} \\
\cmidrule(lr){2-4} \cmidrule(lr){5-7} \cmidrule(lr){8-10} \cmidrule(lr){11-13} \cmidrule(lr){14-16}
& Obj. & Gap & Time & Obj. & Gap & Time & Obj. & Gap & Time & Obj. & Gap & Time & Obj. & Gap & Time \\
\midrule
PyVRP & 3.48 & 0.00\% & 50.56 & 2.54 & 0.00\% & 39.55 & 3.45 & 0.00\% & 33.09 & 3.69 & 0.00\% & 43.02 & 5.95 & 0.00\% & 35.56 \\
OR-Tools & 3.52 & 1.15\% & 50.00 & 2.54 & 0.00\% & 50.00 & 3.49 & 1.16\% & 50.00 & 3.74 & 1.36\% & 50.00 & 6.01 & 1.01\% & 50.00 \\
vrp-cli & 3.61 & 3.74\% & 84.11 & 2.61 & 2.76\% & 81.53 & 3.62 & 4.93\% & 82.16 & 3.80 & 2.98\% & 79.81 & 6.04 & 1.51\% & 65.22 \\
RF-POMO & 3.81 & 9.48\% & 0.15 & 2.79 & 9.84\% & 0.16 & 3.88 & 12.46\% & 0.16 & 4.03 & 9.21\% & 0.16 & 6.32 & 6.22\% & 0.17 \\
RF-MVMoE & 3.74 & 7.47\% & 0.24 & 2.76 & 8.66\% & 0.28 & 3.82 & 10.72\% & 0.28 & 3.95 & 7.05\% & 0.24 & 6.29 & 5.71\% & 0.27 \\
RF-TE & 3.78 & 8.62\% & 0.15 & 2.79 & 9.84\% & 0.15 & 3.85 & 11.59\% & 0.15 & 4.01 & 8.67\% & 0.16 & 6.29 & 5.71\% & 0.17 \\
HF-DRL & 3.87 & 11.21\% & 0.48 & 2.87 & 12.99\% & 0.54 & 3.92 & 13.62\% & 0.51 & 4.15 & 12.47\% & 0.52 & 6.75 & 13.45\% & 0.52 \\
VaP-AM & 3.81 & 9.48\% & 0.31 & 2.83 & 11.42\% & 0.31 & 3.88 & 12.46\% & 0.30 & 4.09 & 10.84\% & 0.30 & 6.67 & 12.10\% & 0.36 \\
VaP-POMO & 3.61 & 3.74\% & 3.04 & 2.65 & 4.33\% & 3.05 & 3.65 & 5.80\% & 3.01 & 3.83 & 3.79\% & 3.09 & 6.30 & 5.88\% & 3.58 \\
\textbf{VaP-CSMV} & \textbf{3.58} & \textbf{2.87\%} & 1.13 & \textbf{2.60} & \textbf{2.36\%} & 1.14 & \textbf{3.54} & \textbf{2.61\%} & 1.08 & \textbf{3.77} & \textbf{2.17\%} & 1.38 & \textbf{6.11} & \textbf{2.69\%} & 1.57 \\
\bottomrule
\end{tabular}
\label{baseline1}
}
\end{table*}

\begin{table*}[t]
\centering
\caption{Detailed Performance Comparison on Datasets for Problem Size \textbf{$N=100, K=30$}.}
\label{tab:baseline_100}
\resizebox{\textwidth}{!}{
\begin{tabular}{lccccccccccccccc}
\toprule
\multirow{2}{*}{Method} & \multicolumn{3}{c}{HFCVRP} & \multicolumn{3}{c}{HFOVRP} & \multicolumn{3}{c}{HFVRPB} & \multicolumn{3}{c}{HFVRPL} & \multicolumn{3}{c}{HFVRPTW} \\
\cmidrule(lr){2-4} \cmidrule(lr){5-7} \cmidrule(lr){8-10} \cmidrule(lr){11-13} \cmidrule(lr){14-16}
& Obj. & Gap & Time & Obj. & Gap & Time & Obj. & Gap & Time & Obj. & Gap & Time & Obj. & Gap & Time \\
\midrule
PyVRP & 5.64 & 0.00\% & 236.59 & 4.08 & 0.00\% & 158.70 & 5.51 & 0.00\% & 149.85 & 5.83 & 0.00\% & 194.63 & 9.37 & 0.00\% & 137.86 \\
OR-Tools & 5.85 & 3.72\% & 200.00 & 4.16 & 1.96\% & 200.00 & 5.72 & 3.81\% & 200.00 & 6.13 & 5.15\% & 200.00 & 9.64 & 2.88\% & 200.00 \\
vrp-cli & 6.28 & 11.35\% & 355.99 & 4.35 & 6.62\% & 291.65 & 6.13 & 11.25\% & 288.67 & 6.30 & 8.06\% & 279.89 & 9.76 & 4.16\% & 155.23 \\
RF-POMO & 6.69 & 18.62\% & 0.31 & 4.61 & 12.99\% & 0.31 & 6.41 & 16.33\% & 0.31 & 6.83 & 17.15\% & 0.31 & 10.22 & 9.07\% & 0.33 \\
RF-MVMoE & 6.43 & 14.01\% & 0.38 & 4.45 & 9.07\% & 0.28 & 6.15 & 11.62\% & 0.28 & 6.53 & 12.01\% & 0.24 & 10.20 & 8.86\% & 0.27 \\
RF-TE & 6.54 & 15.96\% & 0.30 & 4.62 & 13.24\% & 0.30 & 6.28 & 13.97\% & 0.31 & 6.67 & 14.41\% & 0.31 & 10.21 & 8.96\% & 0.32 \\
HF-DRL & 6.68 & 18.44\% & 0.95 & 4.84 & 18.63\% & 1.05 & 6.55 & 18.87\% & 1.02 & 6.81 & 16.81\% & 1.08 & 11.03 & 17.72\% & 1.21 \\
VaP-AM & 6.57 & 16.49\% & 0.68 & 4.74 & 16.18\% & 0.68 & 6.40 & 16.15\% & 0.62 & 6.74 & 15.61\% & 0.61 & 10.93 & 16.65\% & 0.71 \\
VaP-POMO & 6.16 & 9.22\% & 4.99 & 4.42 & 8.33\% & 4.66 & 6.05 & 9.80\% & 4.47 & 6.34 & 8.75\% & 4.51 & 10.06 & 7.36\% & 5.03 \\
\textbf{VaP-CSMV} & \textbf{5.95} & \textbf{5.50\%} & 5.03 & \textbf{4.29} & \textbf{5.15\%} & 5.06 & \textbf{5.77} & \textbf{4.72\%} & 5.00 & \textbf{6.09} & \textbf{4.46\%} & 5.12 & \textbf{9.87} & \textbf{5.34\%} & 5.72 \\
\bottomrule
\end{tabular}
\label{baseline2}
}
\end{table*}

\section{Experiments and Analysis}
In this section, we conduct comprehensive numerical experiments and performance analyses. The proposed model was implemented on an Ubuntu 24.04 Linux server equipped with dual AMD EPYC 9554 64-Core Processors and 768 GB of RAM. Training was performed on an NVIDIA L40S GPU (48 GB), with each model requiring between 9 and 15 hours for convergence.

%All training runs are conducted on NVIDIA L40S GPU with 48 GB of memory and take between 9 and 15 hours per model. 
% All evaluations were conducted on a Red Hat Enterprise Linux server equipped with an Intel(R) Xeon(R) Gold 6248R CPU @ 3.00 GHz and an NVIDIA GeForce RTX 3090 GPU. 
% \textcolor{green}{The source code is publicly available to ensure reproducibility at the GitHub link.}

The model \revmh{projects} all node states and problem prompts into a unified embedding space with a hidden dimension of $d_h = 128$. The cross-semantic encoder comprises $U = 6$ stacked Transformer sub-layers, employing a multi-head attention (MHA) mechanism with $n_{head} = 8$. Synthetic instances for training and testing are generated online using an established generator~\cite{berto2024routefinder}, specifically configured to incorporate heterogeneous fleet characteristics. \revmh{The model is trained independently on datasets with customer scales $N=50$ and $N=100$, paired with fleet sizes of $K=20$ and $K=30$, respectively.} Training spans a maximum of 1000 epochs, with each epoch consisting of 40 batches of 250 instances. Model performance is assessed using a fixed validation set and an early stopping mechanism with a defined tolerance threshold to mitigate overfitting. %We optimize the network using a cosine annealing learning rate schedule, decaying from an initial $3 \times 10^{-4}$ to $2 \times 10^{-4}$ after 20 warmup iterations, alongside a weight decay coefficient of 0.01. 
Network optimization utilizes a cosine annealing learning rate schedule—decaying from $3 \times 10^{-4}$ to $2 \times 10^{-4}$ after 20 warmup iterations—and a weight decay coefficient of 0.01. 
The initial entropy regularization coefficient is 0.03, with decay initiated after $40\%$ of the training process. Finally, to ensure numerical stability during training, the covariance truncation range is bounded to $[0.1, 5.0]$, complemented by a gradient detach probability of 0.15.

\subsection{Comparative Analysis}
We evaluate the proposed  VaP-CSMV framework against a comprehensive suite of baseline methods, categorized into traditional heuristic solvers and DRL-based approaches. The traditional heuristic solvers comprise: \textbf{PyVRP}~\cite{wouda2024PyVRP}, a high-performance solver utilizing a hybrid genetic search~\cite{vidal2022hybrid} to provide near-optimal reference solutions; \textbf{OR-Tools}~\cite{ortools}, a widely adopted combinatorial optimization toolkit from Google; and \textbf{vrp-cli}~\cite{builuk_rosomaxa_2023}, a lightweight open-source solver that employs the Adaptive Large Neighborhood Search (ALNS) algorithm. These traditional solvers are executed on a single CPU core to ensure a consistent performance baseline. For PyVRP and vrp-cli, the maximum number of iterations is restricted to $100 \times N$, where $N$ denotes the node scale. Since OR-Tools does not permit direct iteration control, its computation time is capped at 50 and 200 seconds for 50-node and 100-node instances, respectively. Regarding DRL-based approaches, given the absence of end-to-end models natively supporting multi-variant HFVRP, we adapt several representative state-of-the-art \revmh{neural solvers}. Specifically, we evaluate three models from the ROUTEFINDER framework~\cite{berto2024routefinder}: \textbf{RF-POMO}~\cite{kwon2020pomo}, which utilizes the POMO encoder architecture~\cite{kwon2020pomo} and multi-start sampling; %adopting the POMO encoder architecture combined with multi-start sampling to enhance training stability; 
\revmh{\textbf{RF-MVMoE}}~\cite{zhou2024mvmoe}, an enhanced variant based on the Mixture-of-Experts (MoE) architecture;
%an enhanced version based on the Mixture-of-Experts architecture within the ROUTEFINDER framework; 
and \textbf{RF-TE}, the framework’s flagship model that achieves state-of-the-art performance across multiple VRP variants. As these models lack the capability to actively select heterogeneous vehicles, their generated routes are paired with the SCIP solver for optimal vehicle assignment, ensuring a fair comparison. We also include \textbf{HF-DRL}~\cite{li2022deep}, \revmh{a model originally designed for the heterogeneous capacitated VRP (HCVRP)}. To address its original limitations regarding complex constraints, we integrated the core architecture of HF-DRL into our multi-variant environment for retraining and evaluation. \revmh{Additionally, we evaluate \textbf{VaP-AM} and \textbf{VaP-POMO}, two ablation baselines that combine our VaP mechanism with the standard Attention Model~\cite{kool2018attention} and the POMO architecture~\cite{kwon2020pomo}, respectively.} For inference, the neural solvers based on the ROUTEFINDER framework follow their native decoding strategies, while all other DRL-based methods employ random sampling with 1280 samples.

The proposed VaP-CSMV was trained on five fundamental VRP variants and evaluated across a test set of 1000 instances for each problem scale and configuration. Comprehensive performance metrics are summarized in Table~\ref{tab:baseline_50} and Table~\ref{tab:baseline_100}. As indicated by the results in Table~\ref{baseline1}, VaP-CSMV maintains an optimality gap within 3\% for 50-node instances, outperforming existing \revmh{neural solvers}. Notably, the solution quality of VaP-CSMV closely approaches that of traditional solvers, while its inference time is orders of magnitude lower than their total solution time. For the 100-node scale instances presented in Table~\ref{baseline2}, VaP-CSMV maintains an optimality gap within 6\%, consistently surpassing existing state-of-the-art \revmh{neural solvers}. Furthermore, its performance is comparable or occasionally superior to traditional solvers, offering significantly higher computational efficiency. These results substantiate the robust potential and practical efficacy of the proposed VaP-CSMV framework.

%To compensate for the limitation of its native environment in handling constraints, we adapted its core algorithm to our multi-variant environment for retraining and evaluation. And \textbf{VaP-AM}, an ablation baseline integrating our VaP paradigm with the standard Attention Model~\cite{kool2018attention}. During inference, the RF series employs its native decoding strategies, while all other DRL methods utilize random sampling with $1280$ samples. 

%Meanwhile, its solution quality closely approaches that of traditional solvers, but with a drastically reduced inference time. 

\begin{table*}[!t]
\centering
\caption{Zero-shot generalization performance on various HFVRP variants.}
\resizebox{\textwidth}{!}{
\begin{tabular}{l *{15}{c}}
\toprule
\multirow{2}{*}{Method} & \multicolumn{3}{c}{HFOVRPB} & \multicolumn{3}{c}{HFOVRPL} & \multicolumn{3}{c}{HFOVRPTW} & \multicolumn{3}{c}{HFVRPBL} & \multicolumn{3}{c}{HFVRPBTW} \\
\cmidrule(lr){2-4} \cmidrule(lr){5-7} \cmidrule(lr){8-10} \cmidrule(lr){11-13} \cmidrule(lr){14-16}
& Obj. & Gap & Time & Obj. & Gap & Time & Obj. & Gap & Time & Obj. & Gap & Time & Obj. & Gap & Time \\
\midrule
PyVRP & 2.74 & 0.00\% & 30.95 & 2.54 & 0.00\% & 37.44 & 4.33 & 0.00\% & 28.57 & 3.80 & 0.00\% & 29.41 & 6.96 & 0.00\% & 30.94 \\
OR-Tools & 2.75 & 0.36\% & 50.00 & 2.54 & 0.00\% & 50.00 & 4.34 & 0.23\% & 50.00 & 3.88 & 2.11\% & 50.00 & 7.02 & 0.86\% & 50.00 \\
vrp-cli & 2.82 & 2.92\% & 85.83 & 2.60 & 2.36\% & 91.88 & 4.38 & 1.15\% & 72.32 & 3.96 & 4.21\% & 84.80 & 7.03 & 1.01\% & 70.70 \\
RF-POMO & 3.03 & 10.58\% & 0.16 & 2.79 & 9.84\% & 0.16 & 4.62 & 6.70\% & 0.17 & 4.26 & 12.11\% & 0.16 & 7.35 & 5.60\% & 0.18 \\
RF-MVMoE & 3.01 & 9.85\% & 0.22 & 2.76 & 8.66\% & 0.22 & 4.60 & 6.24\% & 0.32 & 4.23 & 11.32\% & 0.29 & 7.28 & 4.60\% & 0.27 \\
RF-TE & 3.02 & 10.22\% & 0.16 & 2.79 & 9.84\% & 0.16 & 4.61 & 6.47\% & 0.22 & 4.28 & 12.63\% & 0.21 & 7.28 & 4.60\% & 0.18 \\
VaP-POMO & 2.97 & 8.39\% & 0.79 & 2.71 & 6.69\% & 0.72 & 4.74 & 9.47\% & 0.52 & 4.13 & 8.68\% & 0.78 & 7.43 & 6.75\% & 0.93 \\
VaP-CSMV & \textbf{2.88} & \textbf{5.11\%} & 1.49 & \textbf{2.71} & \textbf{6.69\%} & 1.40 & \textbf{4.56} & \textbf{5.31\%} & 1.31 & \textbf{4.00} & \textbf{5.26\%} & 1.20 & \textbf{7.19} & \textbf{3.30\%} & 1.33 \\
\bottomrule
\end{tabular}
}

\vspace{1em}

\resizebox{\textwidth}{!}{
\begin{tabular}{l *{15}{c}}
\toprule
\multirow{2}{*}{Method} & \multicolumn{3}{c}{HFVRPLTW} & \multicolumn{3}{c}{HFOVRPBTW} & \multicolumn{3}{c}{HFOVRPLTW} & \multicolumn{3}{c}{HFOVRPBL} & \multicolumn{3}{c}{HFOVRPBLTW} \\
\cmidrule(lr){2-4} \cmidrule(lr){5-7} \cmidrule(lr){8-10} \cmidrule(lr){11-13} \cmidrule(lr){14-16}
& Obj. & Gap & Time & Obj. & Gap & Time & Obj. & Gap & Time & Obj. & Gap & Time & Obj. & Gap & Time \\
\midrule
PyVRP & 6.14 & 0.00\% & 33.52 & 4.94 & 0.00\% & 27.28 & 4.33 & 0.00\% & 28.12 & 2.74 & 0.00\% & 28.26 & 4.94 & 0.00\% & 27.44 \\
OR-Tools & 6.23 & 1.47\% & 50.00 & 4.95 & 0.20\% & 50.00 & 4.35 & 0.46\% & 50.00 & 2.76 & 0.73\% & 50.00 & 4.95 & 0.20\% & 50.00 \\
vrp-cli & 6.23 & 1.47\% & 69.13 & 4.99 & 1.01\% & 66.81 & 4.38 & 1.15\% & 65.47 & 2.82 & 2.92\% & 85.41 & 4.99 & 1.01\% & 64.08 \\
RF-POMO & 6.51 & 6.03\% & 0.17 & 5.24 & 6.07\% & 0.18 & 4.61 & 6.47\% & 0.17 & 3.04 & 10.95\% & 0.16 & 5.24 & 6.07\% & 0.18 \\
RF-MVMoE & 6.49 & 5.70\% & 0.23 & 5.21 & 5.47\% & 0.25 & 4.61 & 6.47\% & 0.24 & 3.02 & 10.22\% & 0.22 & 5.21 & 5.47\% & 0.25 \\
RF-TE & 6.50 & 5.86\% & 0.17 & 5.21 & 5.47\% & 0.18 & 4.61 & 6.47\% & 0.17 & 3.03 & 10.58\% & 0.16 & 5.21 & 5.47\% & 0.18 \\
VaP-POMO & 6.55 & 6.68\% & 0.89 & 5.41 & 9.51\% & 0.76 & 4.74 & 9.47\% & 0.85 & 2.97 & 8.39\% & 0.77 & 5.45 & 10.32\% & 0.89 \\
VaP-CSMV & \textbf{6.32} & \textbf{2.93\%} & 1.28 & \textbf{5.19} & \textbf{5.06\%} & 1.40 & \textbf{4.55} & \textbf{5.08\%} & 1.30 & \textbf{2.89} & \textbf{5.47\%} & 1.16 & \textbf{5.19} & \textbf{5.06\%} & 1.33 \\
\bottomrule
\end{tabular}
}
\label{variants_generalization}
\end{table*}

\subsection{Ablation Study}
\begin{figure}[!t]
\centering
\includegraphics[width=0.8\columnwidth]{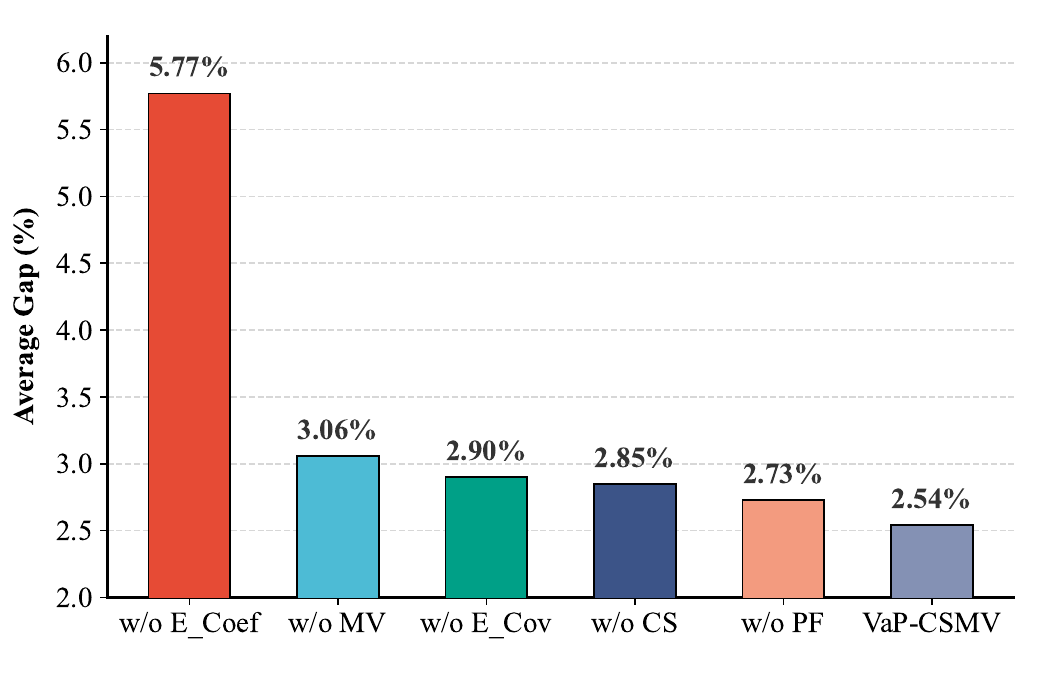}
\caption{Assemble Analysis of The VaP-CSMV Framework.}
\label{fig_assemble_analysis}
\end{figure}

To evaluate the individual contribution of each core component within the VaP-CSMV framework, we compare the full model against several \revmh{ablated variants}. Specifically, \textbf{w/o CS} denotes the removal of the cross-semantic encoder, where the model relies on a conventional Transformer encoder; \textbf{w/o MV} indicates the replacement of the multi-view decoder with a standard Transformer decoder; \textbf{w/o PF} involves excluding problem-specific features from the input layer, requiring the model to infer parameter information without explicit constraint prompts; \textbf{w/o E\_Coef} represents the omission of entropy coefficient regularization during the calculation of the training loss; and \textbf{w/o E\_Cov} indicates the omission of the entropy covariance mechanism from the training control process. The experimental results are illustrated in Figure \ref{fig_assemble_analysis}, with components ranked in descending order of their impact on performance degradation.

The experimental results demonstrate that the full \textbf{VaP-CSMV} model achieves the highest performance, yielding an average optimality gap of 2.54\%. Compared to the ablated variants, the full model exhibits a significant performance advantage, underscoring the \revmh{necessity of integrating all proposed components.} Among the ablated models, the \textbf{w/o E\_Coef} variant experiences the most substantial performance degradation, with the average gap rising to 5.77\%. This indicates that within the VaP mechanism, entropy regularization serves as a critical mechanism for maintaining policy exploration and preventing the model from converging to local optima. Furthermore, the performance drop (2.90\%) observed in the \textbf{w/o E\_Cov} variant substantiates the importance of enhanced exploration during training. The removal of the multi-view decoder or the cross-semantic encoder leads to an increased gap, suggesting that single-view architectures are insufficient for capturing the complex multi-semantic information inherent in the VaP mechanism; whereas the multi-view mechanism effectively integrates features across different dimensions to enhance solution quality. Additionally, excluding problem-specific features (\textbf{w/o PF}) increases the gap to 2.73\%. Although this impact is relatively minor, it demonstrates that incorporating problem prompts effectively assists the model in fine-tuning policies for specific variant configurations.

\begin{table*}[!t]
\centering
\caption{Detailed Performance Comparison on Datasets for Problem Size $N = 80, K = 30$.}
\label{tab:hfvrp_scale_80}
\resizebox{\textwidth}{!}{
\begin{tabular}{l *{15}{c}}
\toprule
\multirow{2}{*}{Method} & \multicolumn{3}{c}{HFCVRP} & \multicolumn{3}{c}{HFOVRP} & \multicolumn{3}{c}{HFVRPB} & \multicolumn{3}{c}{HFVRPL} & \multicolumn{3}{c}{HFVRPTW} \\
\cmidrule(lr){2-4} \cmidrule(lr){5-7} \cmidrule(lr){8-10} \cmidrule(lr){11-13} \cmidrule(lr){14-16}
& Obj. & Gap & Time & Obj. & Gap & Time & Obj. & Gap & Time & Obj. & Gap & Time & Obj. & Gap & Time \\
\midrule
PyVRP & 4.82 & 0.00\% & 131.99 & 3.48 & 0.00\% & 93.51 & 4.71 & 0.00\% & 85.22 & 5.01 & 0.00\% & 107.77 & 8.00 & 0.00\% & 83.10 \\
OR-Tools & 5.00 & 3.73\% & 100.00 & 3.53 & 1.44\% & 100.00 & 4.86 & 3.18\% & 100.00 & 5.24 & 4.59\% & 100.00 & 8.21 & 2.63\% & 100.00 \\
vrp-cli & 5.23 & 8.51\% & 249.67 & 3.68 & 5.75\% & 199.46 & 5.14 & 9.13\% & 218.63 & 5.32 & 6.19\% & 209.23 & 8.29 & 3.62\% & 148.62 \\
RF-POMO & 5.49 & 13.90\% & 0.25 & 3.85 & 10.63\% & 0.25 & 5.35 & 13.59\% & 0.25 & 5.62 & 12.18\% & 0.25 & 8.69 & 8.62\% & 0.27 \\
RF-MVMoE & 5.42 & 12.45\% & 0.33 & 3.81 & 9.48\% & 0.34 & 5.28 & 12.10\% & 0.34 & 5.52 & 10.18\% & 0.33 & 8.68 & 8.50\% & 0.36 \\
RF-TE & 5.45 & 13.07\% & 0.24 & 3.84 & 10.34\% & 0.24 & 5.29 & 12.31\% & 0.24 & 5.59 & 11.58\% & 0.24 & 8.67 & 8.38\% & 0.26 \\
HF-DRL & 5.55 & 15.15\% & 0.64 & 3.84 & 10.34\% & 0.65 & 5.29 & 12.31\% & 0.68 & 5.68 & 13.37\% & 0.62 & 9.28 & 16.00\% & 0.82 \\
VaP-AM & 5.51 & 14.32\% & 0.51 & 3.93 & 12.93\% & 0.52 & 5.41 & 14.86\% & 0.50 & 5.69 & 13.57\% & 0.54 & 9.26 & 15.75\% & 0.63 \\
VaP-POMO & 5.23 & 8.51\% & 2.61 & 3.71 & 6.61\% & 2.62 & 5.15 & 9.34\% & 2.42 & 5.41 & 7.98\% & 2.54 & 8.72 & 9.00\% & 2.68 \\
\textbf{VaP-CSMV} & \textbf{5.23} & \textbf{8.51\%} & 3.05 & \textbf{3.68} & \textbf{5.75\%} & 3.07 & \textbf{5.04} & \textbf{7.01\%} & 2.85 & \textbf{5.35} & \textbf{6.79\%} & 2.94 & \textbf{8.39} & \textbf{4.88\%} & 3.28 \\
\bottomrule
\end{tabular}
}
\end{table*}

\begin{table*}[!t]
\centering
\caption{Detailed Performance Comparison on Datasets for Problem Size $N = 120, K = 40$.}
\label{tab:hfvrp_scale_120}
\resizebox{\textwidth}{!}{
\begin{tabular}{l *{15}{c}}
\toprule
\multirow{2}{*}{Method} & \multicolumn{3}{c}{HFCVRP} & \multicolumn{3}{c}{HFOVRP} & \multicolumn{3}{c}{HFVRPB} & \multicolumn{3}{c}{HFVRPL} & \multicolumn{3}{c}{HFVRPTW} \\
\cmidrule(lr){2-4} \cmidrule(lr){5-7} \cmidrule(lr){8-10} \cmidrule(lr){11-13} \cmidrule(lr){14-16}
& Obj. & Gap & Time & Obj. & Gap & Time & Obj. & Gap & Time & Obj. & Gap & Time & Obj. & Gap & Time \\
\midrule
PyVRP & 6.43 & 0.00\% & 312.67 & 4.63 & 0.00\% & 218.23 & 6.26 & 0.00\% & 204.86 & 6.61 & 0.00\% & 260.17 & 10.47 & 0.00\% & 183.31 \\
OR-Tools & 6.84 & 6.38\% & 250.00 & 4.79 & 3.46\% & 250.00 & 6.61 & 5.59\% & 250.00 & 7.17 & 8.47\% & 250.00 & 10.89 & 4.01\% & 250.00 \\
vrp-cli & 7.33 & 14.00\% & 560.22 & 5.04 & 8.86\% & 384.60 & 7.15 & 14.22\% & 404.33 & 7.34 & 11.04\% & 354.64 & 11.02 & 5.25\% & 195.48 \\
RF-POMO & 7.78 & 21.00\% & 0.49 & 5.28 & 14.04\% & 0.49 & 7.36 & 17.57\% & 0.49 & 7.89 & 19.36\% & 0.50 & 11.56 & 10.41\% & 0.51 \\
RF-MVMoE & 7.49 & 16.49\% & 0.65 & 5.11 & 10.37\% & 0.58 & 7.08 & 13.10\% & 0.51 & 7.55 & 14.22\% & 0.51 & 11.54 & 10.22\% & 0.54 \\
RF-TE & 7.59 & 18.04\% & 0.42 & 5.27 & 13.82\% & 0.49 & 7.18 & 14.70\% & 0.38 & 7.68 & 16.19\% & 0.37 & 11.54 & 10.22\% & 0.50 \\
HF-DRL & 7.68 & 19.44\% & 1.05 & 5.52 & 19.22\% & 1.02 & 7.39 & 18.05\% & 1.01 & 7.85 & 18.76\% & 1.10 & 12.05 & 15.09\% & 1.22 \\
VaP-AM & 7.63 & 18.66\% & 0.81 & 5.41 & 16.85\% & 0.79 & 7.34 & 17.25\% & 0.82 & 7.76 & 17.40\% & 0.82 & 12.04 & 15.00\% & 0.91 \\
VaP-POMO & 7.16 & 11.35\% & 4.22 & 5.08 & 9.72\% & 4.16 & 6.95 & 11.02\% & 4.04 & 7.32 & 10.74\% & 4.08 & 11.60 & 10.79\% & 4.68 \\
\textbf{VaP-CSMV} & \textbf{7.05} & \textbf{9.64\%} & 9.35 & \textbf{4.95} & \textbf{6.91\%} & 8.77 & \textbf{6.77} & \textbf{8.15\%} & 8.34 & \textbf{7.21} & \textbf{9.08\%} & 8.48 & \textbf{11.15} & \textbf{6.49\%} & 9.15 \\
\bottomrule
\end{tabular}
}
\end{table*}

\subsection{Impact of Vehicle Heterogeneity}
\begin{figure}[!t]
\centering
\includegraphics[width=0.9\columnwidth]{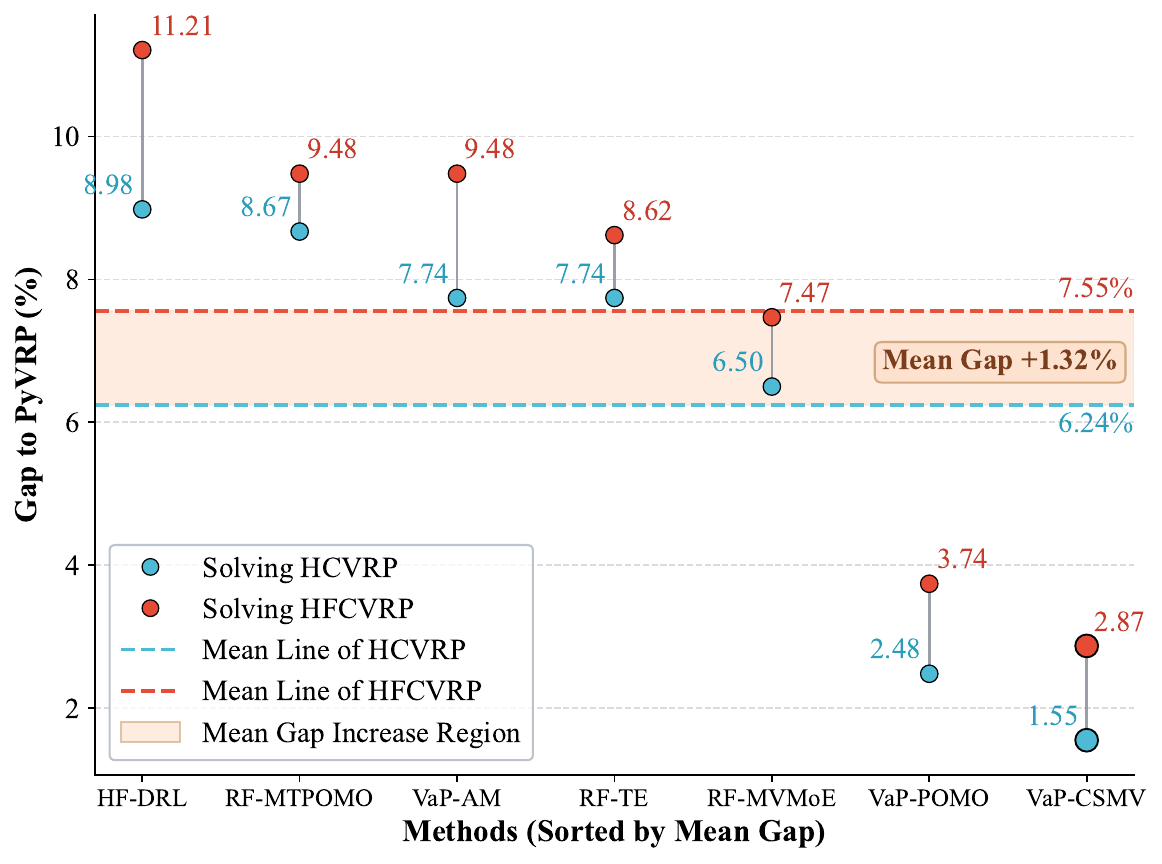}
\caption{Gap to PyVRP (\%) of different neural solvers on HCVRP and HFCVRP. }
\label{fig_hf_analysis}
\end{figure}

\revmh{To validate the necessity of modeling multi-dimensional vehicle heterogeneity and the effectiveness of our method, we compare several neural solvers across two settings: HCVRP and HFCVRP, using PyVRP as the reference baseline. Specifically, HCVRP accounts for vehicle capacity heterogeneity and variable travel costs, whereas HFCVRP further incorporates fixed cost heterogeneity, thereby offering a more realistic and challenging setting. As shown in Figure \ref{fig_hf_analysis}, the average \wsj{gap} of the evaluated neural solvers increases from 6.24\% on HCVRP to 7.55\% on HFCVRP, yielding an average increase of 1.31 percentage points. This indicates that when fixed costs are introduced, vehicle dispatching becomes more critical, and the coupling between vehicle selection and route construction strengthens, making the problem substantially more difficult. These results confirm that modeling heterogeneity in capacity alone is insufficient for practical HFCVRP applications, highlighting the need to capture multi-dimensional vehicle attributes. In contrast, the proposed VaP-CSMV achieves the lowest gap in both settings, recording 1.55\% on HCVRP and 2.87\% on HFCVRP. Despite the increased difficulty of HFCVRP, our method consistently maintains superior solution quality, demonstrating its ability to effectively capture heterogeneous vehicle attributes, namely capacities, variable travel costs, and fixed costs. This underscores the robustness of the proposed \wsj{framework} in handling complex, heterogeneous fleet routing decisions.
}

\subsection{Generalization Analysis}
To evaluate the zero-shot generalization capabilities of the proposed framework, we utilized a test set comprising novel HFVRP variants subject to multiple constraints. Notably, none of these configurations were encountered by VaP-CSMV during its training phase. Detailed experimental results are presented in Table~\ref{variants_generalization}. The results demonstrate that \revmh{VaP-CSMV} consistently maintains an optimality gap within 7\% across all \wsj{variants}. It significantly outperforms existing state-of-the-art neural solvers, such as RF-POMO and RF-MVMoE. Compared to traditional solvers like PyVRP and OR-Tools, VaP-CSMV achieves competitive solution quality while maintaining seconds-level inference times. Even when evaluated on variants with \revmh{additional constraints}%superimposed
—including backhauls, linehauls, open routes, time-windows, and distance limits—the framework maintains stable performance. These results substantiate that VaP-CSMV possesses robust zero-shot generalization capabilities for unseen HFVRP variants and exhibits high adaptability across diverse problem configurations.

To evaluate the model's generalization performance across problem scales, we constructed two large-scale HFVRP test sets that exceed the distribution range of the training set (see Table \ref{tab:hfvrp_scale_80} and Table \ref{tab:hfvrp_scale_120}, where $N$ denotes the node scale and $K$ denotes the fleet size). Neither the problem scales nor the fleet sizes in these instances were encountered during training, providing a rigorous test of the model's generalization capabilities in \revmh{out-of-distribution scenarios}. Experimental results indicate that VaP-CSMV achieves an optimality gap below 9\% on the $(N=80, K=30)$ test set, and maintains a stable gap of approximately 10\% on the larger $(N=120, K=40)$ test set. Its performance significantly surpasses that of the neural solvers based on the ROUTEFINDER framework and the HF-DRL and VaP-AM baselines. These findings substantiate that VaP-CSMV does not merely fit specific scale distributions but possesses robust zero-shot scale generalization capabilities. Compared to the traditional solver PyVRP, the inference time of  VaP-CSMV scales linearly with problem size, consistently remaining under 10 seconds. In contrast, the total solution time required by PyVRP increases significantly as the problem scale expands. This indicates that the VaP-CSMV framework maintains stable solution quality and high computational efficiency on large-scale instances, offering a promising technical pathway for efficiently resolving large-scale VRPs.

\section{Conclusion}
In this paper, we investigated HFVRP subject to a wide array of complex, real-world constraints. We proposed a unified DRL framework driven by the innovative Vehicle-as-Prompt (VaP) mechanism, which reformulates the multi-variant HFVRP into a streamlined, single-stage autoregressive decision process. Specifically, the VaP-CSMV framework utilizes a dual-attention mechanism to accommodate multiple problem variants, \revmh{while leveraging a cross-semantic encoder and a multi-view decoder to capture rich, high-dimensional interactions between vehicle attributes and node features}.

Comprehensive experiments demonstrate that VaP-CSMV consistently outperforms existing learning-based approaches. Compared to state-of-the-art traditional heuristic solvers, the proposed framework achieves competitive solution quality while delivering a significant advantage in computational efficiency. Furthermore, it exhibits robust zero-shot generalization capabilities across unseen HFVRP variants and out-of-distribution problem scales.
Future research may focus on extending the framework to address real-time scheduling under dynamic requests and stochastic traffic conditions, as well as integrating broader physical and environmental constraints into the VaP mechanism. 
%More future work for TRE submission}

%Experiments demonstrate that our method outperforms other learning-based approaches. Compared to state-of-the-art traditional heuristics, it achieves competitive solution quality while significantly reducing computation time. Furthermore, it exhibits robust zero-shot generalization on larger-scale and previously unseen variant instances. Our future work will focus on addressing real-time scheduling with dynamic requests and stochastic traffic, as well as extending the VaP paradigm to incorporate complex physical constraints.

% reference
\bibliographystyle{IEEEtran}
\bibliography{re_mvhfvrp}

\vfill

\end{document}